\documentclass[sigconf]{acmart}

\usepackage{booktabs} 
\usepackage{hyperref}
\usepackage{mathrsfs,mathtools}
\usepackage{amsmath,amssymb}
\usepackage{multirow}
\usepackage{balance}
\usepackage{graphicx}
\usepackage[caption=false]{subfig}

\makeatletter
\def\blfootnote{\gdef\@thefnmark{}\@footnotetext}
\makeatother

\begin{document}

\copyrightyear{2018}
\acmYear{2018}
\setcopyright{acmcopyright}
\acmConference[MM '18]{2018 ACM Multimedia Conference}{October 22--26,
2018}{Seoul, Republic of Korea}
\acmBooktitle{2018 ACM Multimedia Conference (MM '18), October 22--26,
2018, Seoul, Republic of Korea}
\acmPrice{15.00}
\acmDOI{10.1145/3240508.3240533}
\acmISBN{978-1-4503-5665-7/18/10}

\fancyhead{}

\title{Twitter Sentiment Analysis via Bi-sense Emoji Embedding and Attention-based LSTM}

\author{Yuxiao Chen}
\authornote{Yuxiao Chen and Jianbo Yuan contributed equally to this work.}
\affiliation{%
  \institution{Department of Computer Science\\ University of Rochester}
  \streetaddress{250 Hutchison RD}
  \city{Rochester}
  \state{NY}
  \postcode{14627}
}
\email{ychen211@cs.rochester.edu}

\author{Jianbo Yuan}
\affiliation{%
  \institution{Department of Computer Science\\ University of Rochester}
  \streetaddress{250 Hutchison RD}
  \city{Rochester}
  \state{NY}
  \postcode{14627}
}
\email{jyuan10@cs.rochester.edu}
\authornotemark[1]

\author{Quanzeng You}
\affiliation{%
  \institution{Microsoft Research AI}
  \city{Redmond}
  \state{WA}
}
\email{quyou@microsoft.com}

\author{Jiebo Luo}
\affiliation{%
  \institution{Department of Computer Science\\ University of Rochester}
  \streetaddress{250 Hutchison RD}
  \city{Rochester}
  \state{NY}
  \postcode{14627}
}
\email{jluo@cs.rochester.edu}


\begin{abstract}
Sentiment analysis on large-scale social media data is important to bridge the gaps between social media contents and real world activities including political election prediction, individual and public emotional status monitoring and analysis, and so on. Although textual sentiment analysis has been well studied based on platforms such as Twitter and Instagram, analysis of the role of extensive emoji uses in sentiment analysis remains light. In this paper, we propose a novel scheme for Twitter sentiment analysis with extra attention on emojis. We first learn \emph{bi-sense} emoji embeddings under positive and negative sentimental tweets individually, and then train a sentiment classifier by attending on these bi-sense emoji embeddings with an attention-based long short-term memory network (LSTM). Our experiments show that the bi-sense embedding is effective for extracting sentiment-aware embeddings of emojis and outperforms the state-of-the-art models. We also visualize the attentions to show that the bi-sense emoji embedding provides better guidance on the attention mechanism to obtain a more robust understanding of the semantics and sentiments.
\end{abstract}



\maketitle

\section{Introduction}
The rapid growth of social media platforms such as Twitter provides rich multimedia data in large scales for various research opportunities, such as sentiment analysis which focuses on automatically sentiment (positive and negative) prediction on given contents. Sentiment analysis has been widely used in real world applications by analyzing the online user-generated data, such as election prediction, opinion mining and business-related activity analysis. Emojis, which consist of various symbols ranging from cartoon facial expressions to figures such as flags and sports, are widely used in daily communications to express people's feelings \footnote{Real time emoji tracker: http://emojitracker.com/}.  Since their first release in 2010, emojis have taken the place of emoticons (such as ``:-$)$'' and ``:-P'') \cite{pavalanathan2015emoticons} to create a new form of language for social media users \cite{alshenqeeti2016emojis}. According to recent science reports, there are 2,823 emojis in unicode standard in \emph{Emoji 11.0} \footnote{https://emojipedia.org/emoji-11.0/}, with over 50\% of the Instagram posts containing one or more emojis \cite{dimson2015emoji} and 92\% of the online population using emojis \cite{emojiresearch2016}. 

\begin{table*}[h]
	\centering
	\caption{Tweet examples with emojis. The sentiment ground truth is given in the second column. The examples show that inconsistent sentiments exist between emojis and texts.}
	\label{tb:tweets}
	\begin{tabular}{c|c|l}
		\hline
		Emoji & Sentiment & \multicolumn{1}{c}{Tweets} \\ \hline
		\multirow{4}{*}{\includegraphics[height=2em]{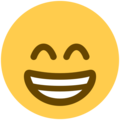}} & \multirow{2}{*}{Positive} & Good morning Justin!!! I hope u have an amazing Friday :) Don't forget to smile \includegraphics[height=0.7em]{54.png}. \\ \cline{3-3} 
		&  & That's awesome :) I'm super keen to hear/see it all \includegraphics[height=0.7em]{54.png}. \\ \cline{2-3} 
		& \multirow{2}{*}{Negative} & I really hate times square personally it's too busy (I'm claustrophobic \includegraphics[height=0.7em]{54.png}). \\ \cline{3-3} 
		&  & Not very good when your sat waiting for your food and there is a queue forming to complain to a manager \includegraphics[height=0.7em]{54.png}. \\ \hline
		\multirow{4}{*}{\includegraphics[height=2em]{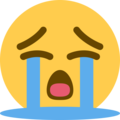}} & \multirow{2}{*}{Positive} & This weather is perfect! \includegraphics[height=0.7em]{22.png} It's just the change I needed. \\ \cline{3-3} 
		&  & The dresses I ordered arrived this morning and they're so pretty \includegraphics[height=0.7em]{22.png}.\\ \cline{2-3} 
		& \multirow{2}{*}{Negative} & Worst headache ever and feel so sick, mum where are you \includegraphics[height=0.7em]{22.png}. \\ \cline{3-3} 
		&  & This nurse always comes mad early in the morning I'm mad tired \includegraphics[height=0.7em]{22.png}. \\ \hline
	\end{tabular}
\end{table*}

The extensive use of emojis has drawn a growing attention from researchers \cite{DBLP:conf/icwsm/LiCHL18,hu2017spice} because the emojis convey fruitful semantical and sentimental information to visually complement the textual information which is significantly useful in understanding the embedded emotional signals in texts \cite{barbieri2016cosmopolitan}. For example, emoji embeddings have been proposed to understand the semantics behind the emojis \cite{EisnerRABR16,li2017joint}, and the embedding vectors can be used to visualize and predict emoji usages given their corresponding contexts. Previous work also shows that, it is useful to pre-train a deep neural network on an emoji prediction task with pre-trained emoji embeddings to learn the emotional signals of emojis for other tasks including sentiment, emotion and sarcasm prediction \cite{felbo2017using}. However, the previous literatures lack in considerations of the linguistic complexities and diversity of emoji. Therefore, previous emoji embedding methods fail to handle the situation when the semantics or sentiments of the learned emoji embeddings contradict the information from the corresponding contexts \cite{hu2017spice}, or when the emojis convey multiple senses of semantics and sentiments such as (\includegraphics[height=0.8em]{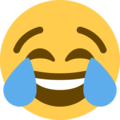} and \includegraphics[height=0.8em]{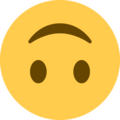}). In practice, emojis can either summarize and emphasis the original tune of their contexts, or express more complex semantics such as irony and sarcasm by being combined with contexts of contradictory semantics or sentiments. For the examples shown in Table \ref{tb:tweets}, the emoji (\includegraphics[height=0.8em]{54.png}) is of consistent sentiment with text to emphasis the sentiment, but is of the opposite sentiment (positive) to the text sentiment (negative) example 3 and 4 to deliver a sense of sarcasm. Conventional emoji analysis can only extract single embedding of each emoji, and such embeddings will confuse the following sentiment analysis model by inconsistent sentiment signals from the input texts and emojis. Moreover, we consider the emoji effect modeling different from the conventional multimodal sentiment analysis which usually includes images and texts in that, image sentiment and text sentiment are usually assumed to be consistent \cite{you2016robust} while it carries no such assumption for texts and emojis.

To tackle such limitations, we propose a novel scheme that consists of an attention-based recurrent neural network (RNN) with robust \emph{bi-sense} emoji embeddings. Inspired by the word sense embedding task in natural language processing (NLP) \cite{iacobacci2015sensembed,li2015multi,song2016sense} where each sense of an ambiguous word responds to one unique embedding vector, the proposed \emph{bi-sense} embedding is a more robust and fine-grained representation of the complicated semantics for emojis where each emoji is embedded into \emph{two distinct vectors}, namely positive-sense and negative-sense vector, respectively. For our specific task which is Twitter sentiment analysis \cite{sarlan2014twitter,kouloumpis2011twitter}, we initialize the bi-sense embedding vectors together with word embedding vectors using word embedding algorithm \emph{fasttext} \cite{bojanowski2016enriching} by extracting two distinct embeddings for each emoji according to the sentiment of its corresponding textual contexts, namely \emph{bi-sense} embedding. A long short-term memory (LSTM) based recurrent neural network is then used for predicting sentiments which is integrated with the pre-trained emoji embedding features by a context-guide and self-selected attention mechanism. 
Because most of the previous Twitter sentiment datasets exclude emojis and there exists little resource that contains sufficient emoji-tweets with sentiment labels, we construct our own emoji-tweets dataset by automatically generating weak labels using a rule-based sentiment analysis algorithm \emph{Vader} \cite{HuttoG14} for pre-traning the networks, and manually labeling a subset of tweets for fine tuning and testing purposes. The experimental results demonstrate that the bi-sense emoji embedding is capable of extracting more distinguished information from emojis and outperforms the state-of-the-art sentiment analysis models with the proposed attention-based LSTM networks. We further visualize the bi-sense emoji embedding to obtain the sentiments and semantics learned by the proposed approach. 

The main contributions of this paper are summarized as follows.
\begin{itemize}
	\vspace{-0.75mm}
	\item We propose a novel bi-sense embedding scheme that learns more robust and fine-grained representations of the complex semantic and sentiment information from emojis.
	\item We propose attention-based LSTM networks to encode both texts and bi-sense emoji embedding which outperform the state-of-the-art sentiment analysis models. The networks can be further extended to tackle tasks with multi-sense embedding inputs.
\end{itemize}

\section{Related Work}

\subsection{Sentiment Analysis}
Sentiment analysis is to extract and quantify subjective information including the status of attitudes, emotions and opinions from a variety of contents such as texts, images and audios \cite{YuanYL15}. Sentiment analysis has been drawing great attentions because of its wide applications in business and government intelligence, political science, sociology and psychology \cite{AlkubaisiKH18,OzturkA18,ElghazalyMH16,Aldahawi15}. From a technical perspective, textual sentiment analysis is first explored by researchers as an NLP task. Methods range from lexical-based approaches using features including keywords \cite{pang2002thumbs,BaccianellaES10} where each word corresponds to a sentiment vector with entries representing the possibility of the word and each sentiment and phase-level features (n-grams and unigrams) \cite{pak2010twitter,wilson2005recognizing}, to deep neural network based embedding approaches including skip-grams, continuous bag-of-words (CBoW) and skip-thoughts \cite{mikolov2013,mikolov2013distributed,bojanowski2016enriching,kiros2015skip}. It was until recent years when researchers start focusing on image and multimodal sentiments \cite{borth2013sentibank,yuan2013sentribute} and analyzing how to take advantage of the cross-modality resources \cite{you2016robust,you2016sentiment}. For multimodal sentiment analysis, an underlying assumption is that both modalities express similar sentiment and such similarity is enforced in order to train a robust sentiment inference model \cite{you2016robust}. However, the same assumption does not stand in modeling textual tweets and emojis because the complexities of natural language exist extensively, such as the use of irony, jokes, sarcasm, etc. \cite{felbo2017using}. 

\subsection{Emojis and Sentiment Analysis}
With the overwhelming development of Internet of Things (IOT), the growing accessibility and popularity of subjective contents have provided new opportunities and challenges for sentiment analysis \cite{pang2008opinion}. For example, social medias such as Twitter and Instagram have been explored because the massive user-generated contents with rich user sentiments \cite{pak2010twitter,agarwal2011sentiment,gonzalez2011identifying} where emojis (and emoticons) are extensively used. Non-verbal cues of sentiment, such as emoticon which is considered as the previous generation of emoji, has been studied for their sentiment effect before emojis take over \cite{Zhao2012,LiuLG12,Hogenboom2013}. For instance, \cite{Zhao2012,Hogenboom2013} pre-define sentiment labels to emoticons and construct a emoticon-sentiment dictionary. \cite{LiuLG12} applies emoticons for smoothing noisy sentiment labels. Similar work from \cite{novak2015sentiment} first considers emoji as a component in extracting the lexical feature for further sentiment analysis. \cite{NovakSSM15} constructs an emoji sentiment ranking based on the occurrences of emojis and the human-annotated sentiments of the corresponding tweets where each emoji is assigned with a sentiment score from negative to positive \footnote{http://kt.ijs.si/data/Emoji\_sentiment\_ranking/}, similar to the SentiWordNet \cite{Esuli006}. However, the relatively intuitive use of emojis by lexical- and dictionary-based approaches lacks insightful understanding of the complexed semantics of emojis. Therefore, inspired by the success of word semantic embedding algorithms such as \cite{mikolov2013distributed,bojanowski2016enriching}, \cite{EisnerRABR16} obtains semantic embeddings of each emoji by averaging the words from its descriptions \footnote{http://www.unicode.org/emoji/charts/} and shows it is effective to take advantage of the emoji embedding for the task of Twitter sentiment analysis. \cite{li2017joint} proposes a convoluntional neural network to predict the emoji occurrence and jointly learns the emoji embedding via a matching layer based on cosine similarities. Despite the growing popularity of Twitter sentiment analysis, there is a limited number of emoji datasets with sentiment labels available because previous studies usually filter out urls, emojis and sometimes emoticons. However, \cite{felbo2017using} shows that it is effective to extract sentiment information from emojis for emotion classification and sarcasm detection tasks in the absence of learning vector-based emoji representations by pre-training a deep neural network to predict the emoji occurrence.

\section{Methodology}
\begin{figure}[h]
	\includegraphics[width=0.5\textwidth]{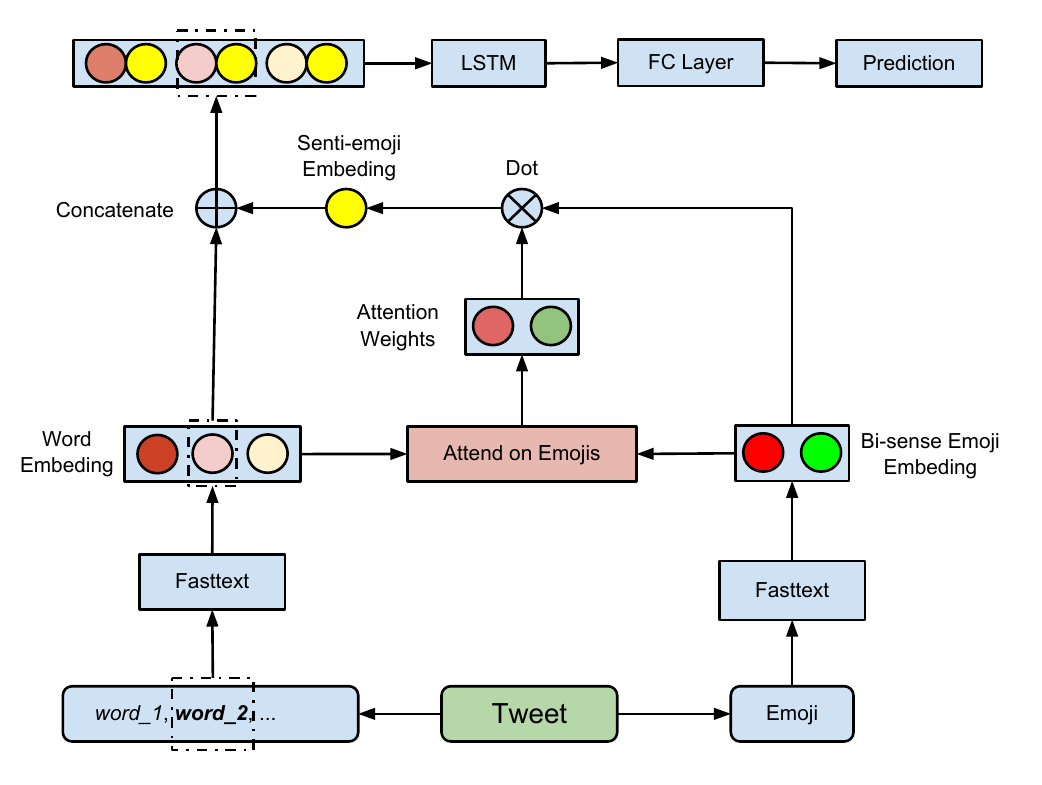}
	\caption{Sentiment analysis via bi-sense emoji embedding and attention-based LSTM network (WATT-BiE-LSTM).}
	\label{fig:method:frame}
\end{figure}

\begin{figure}[t]
	\includegraphics[width=0.48\textwidth]{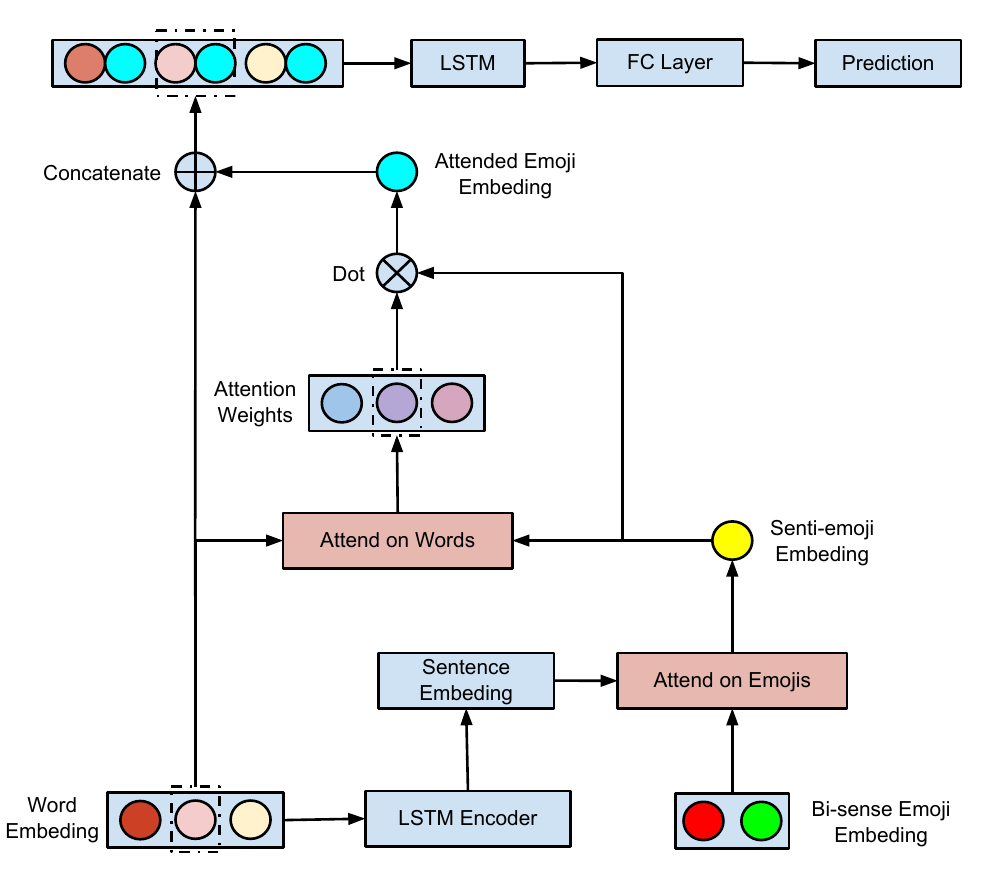}
	\caption{Multi-level attention-based LSTM with bi-sense emoji embedding (MATT-BiE-LSTM).}
	\label{fig:method:frame2}
\end{figure}

We propose two mechanisms, namely Word-guide Attention-based LSTM and Multi-level Attention-based LSTM, to take advantage of bi-sense emoji embedding for the sentiment analysis task. The frameworks of these two methods are shown in Figure \ref{fig:method:frame} and Figure \ref{fig:method:frame2}, respectively. Our workflow includes the following steps: initialization of bi-sense emoji embedding, generating senti-emoji embedding based on self-selected attention, and sentiment classification via the proposed attention-based LSTM networks.

\subsection{Bi-sense Embedding}
Recent research shows great success in word embedding task such as \emph{word2vec} and \emph{fasttext} \cite{mikolov2013,bojanowski2016enriching}. We use \emph{fasttext} to initialize emoji embeddings by considering each emoji as a special word, together with word embeddings. The catch is, different from conventional approaches where each emoji responds to \emph{one} embedding vector (as we call \textbf{word-emoji embedding}), we embed each emoji into two distinct vectors (\textbf{bi-sense emoji embedding}): we first assign two distinct tokens to each emoji, of which one is for the particular emoji used in positive sentimental contexts and the other one is for this emoji used in negative sentimental contexts (text sentiment initialized by \emph{Vader} \cite{HuttoG14}, details will be discussed in Section \ref{sec:exp:data}), respectively; the same \emph{fasttext} training process is used to embed each token into a distinct vector, and we thus obtain the positive-sense and negative-sense embeddings for each emoji.

The \emph{word2vec} is based on the \emph{skip-gram} model whose objective is to maximize the log likelihood calculated by summing the probabilities of current word occurrences given a set of the surrounding words. The \emph{fasttext} model is different by formatting the problem as a binary classification task to predict the occurrence of each context word, with negative samples being randomly selected from the absent context words. Given an input word sequence $\left\{w_{1},w_{2},...,w_{T}\right \}$, and the context word set $W_{c_t}$ and the set of negative word samples $W_{n_t}$ of the current word $w_{t}$, the objective function is obtained based on binary logistic loss as in Equation \ref{eq:obj:fast}:

\begin{equation}
\label{eq:obj:fast}
\sum_{t=1}^{T}\left[\sum_{w_c \in W_{c_t}} \mathcal{L}\left(s\left(w_t,w_c \right) \right)+ \sum_{w_n \in W_{n_t}}\mathcal{L}\left(s\left(w_t,w_n \right ) \right)\right]
\end{equation}

\noindent where $\mathcal{L}(s(\cdot,\cdot))$ denotes the logistic loss of the score function $s(\cdot\ ,\cdot)$ which is computed by summing up the scalar products between the n-gram embeddings of the current word and the context word embedding which is different from \emph{word2vec} where the score is the scalar product between the current word and the context word embedding. We select \emph{fasttext} over \emph{word2vec} mainly because its computational efficiency. In general, the two models yield competitive performances and the comparison between word embeddings is beyond our discussion. Therefore we only show the results derived by the \emph{fasttext} initialization within the scope of this work.

\subsection{Word-guide Attention-based LSTM}
\label{sec:attlstm}
Long short-term memory (LSTM) units have been extensively used to encode textual contents. The basic encoder model consists of a text embedding layer, LSTMs layer, and fully-connected layers for further tasks such as text classifications based on the encoded feature. The operations in an LSTM unit for time step $t$ is formulated in Equation \ref{eq:lstm}: 

\begin{equation}
\label{eq:lstm}
\begin{aligned}
\textbf{i}_t &= \sigma(W_{i}\textbf{x}_{t}+U_{i}\textbf{h}_{t-1}+b_i) \\
\textbf{f}_t &= \sigma(W_{f}\textbf{x}_{t}+U_{f}\textbf{h}_{t-1}+b_f) \\
\textbf{o}_t &= \sigma(W_{o}\textbf{x}_{t}+U_{o}\textbf{h}_{t-1}+b_o) \\
\textbf{g}_t &= \tanh(W_{c}\textbf{x}_{t}+U_{c}\textbf{h}_{t-1}+b_c) \\
\textbf{c}_t &= \textbf{f}_t\odot \textbf{c}_{t-1} + \textbf{i}_t\odot \textbf{g}_t \\
\textbf{h}_t &= \textbf{o}_t\odot\tanh(\textbf{c}_t)
\end{aligned}
\end{equation}

\noindent where $\textbf{h}_{t}$ and $\textbf{h}_{t-1}$ represent the current and previous hidden states, $\textbf{x}_{t}$ denotes the current LSTM input and here we use the embedding $\textbf{w}_{t}$ of the current word $w_{t}$, and $W$ and $U$ denote the weight matrices \cite{hochreiter1997long}. In order to take advantage of the bi-sense emoji embedding, we modify the input layer into the LSTM units. We first obtain the \textbf{senti-emoji embedding} as an weighted average of the bi-sense emoji embedding based on the self-selected attention mechanism. Let $\textbf{e}_{t,i},i\in(1,m)$ represent the $i$-th sense embedding of emoji $e_{t}$ ($m=2$ in our bi-sense embedding), and $f_{att}(\ \cdot\ ,\textbf{w}_t)$ denote the attention function conditioned on the current word embedding, the attention weight $\alpha_i$ and senti-emoji embedding vector $\textbf{v}_t$ is formulated as follows:
\begin{equation}
\label{eq:att}
\begin{aligned}
u_{t,i}&=f_{att}(\textbf{e}_{t,i},\textbf{w}_t)\\
\alpha_{t,i}&=\frac{\exp (u_{t,i})}{\sum_{i=1}^{m} \exp (u_{t,i})}\\
\textbf{v}_{t}&=\sum_{i=1}^{m} \left(\alpha_{t,i}\cdot \textbf{e}_{t,i} \right)
\end{aligned}
\end{equation}

We choose a fully-connected layer with \emph{ReLU} activation as the attention function, and the attention vector $v_{t}$ is concatenated with the word embedding as the new input of the LSTM. Thus the input vector $\textbf{x}_{t}$ in Equation \ref{eq:lstm} becomes $\left[ \textbf{w}_t, \textbf{v}_{t} \right]$. The output of the final LSTM unit is then fed into a fully-connected layer with $sigmoid$ activation to output the tweet sentiment and binary cross-entropy loss is used as the objection function (Equation \ref{eq:obj:senti}) where $N$ is the total number of samples. The motivation behind this model is that each context word guides the attention weights in order to enforce the model to self-select which embedding sense it should attend on. Therefore we denote this model as the Word-guide Attention-based LSTM with Bi-sense emoji embedding (\textbf{WATT-BiE-LSTM}).

\begin{equation}
\label{eq:obj:senti}
\begin{aligned}
\mathcal{L(\theta)}=-\frac{1}{N}\sum_{i=1}^{N}\left(y_{i}\log(p_{i})+(1-y_{i})\log(1-p_{i}) \right)
\end{aligned}
\end{equation}

\subsection{Multi-level Attention-based LSTM}
There is another way of formulating the attention mechanism where the attention weights indicate how the image information (which is emoji in our case) is distributed through the context words as proposed in \cite{wang2016attention,chenCL17}. The modified senti-emoji embedding vector $\textbf{v}$ is thus at the tweet-level instead of the word-level in Equation \ref{eq:att} by replacing the $\textbf{w}_t$ with the final state vector $\textbf{h}$ outputted from the last LSTM unit, as shown in Equation \ref{eq:att2}:

\begin{equation}
\label{eq:att2}
\begin{aligned}
\alpha_{i}&=\frac{\exp \left( f_{att}(\textbf{e}_{i},\textbf{h}) \right)}{\sum_{i=1}^{m} \exp\left (f_{att}(\textbf{e}_{i},\textbf{h})\right)}\\
\textbf{v}'&=\sum_{i=1}^{m} \left(\alpha_{i}\cdot \textbf{e}_{i} \right)
\end{aligned}
\end{equation}

The derived senti-emoji embedding $\textbf{v}'$ is then used to calculate an additional layer of attention following \cite{wang2016attention,chenCL17}. Given the input tweet sequence $\left\{ w_{1},w_{2},...,w_{T}\right\}$, the attention weight $\alpha_{t}',t\in(1,T)$ conditioned on the senti-emoji embedding is formulated as follows:

\begin{equation}
\label{eq:att3}
\begin{aligned}
\alpha_{t}'&=\frac{\exp \left(f_{att}(\textbf{w}_{t},\textbf{v}') \right)}{\sum_{i=1}^{m} \exp\left( f_{att}(\textbf{w}_{t},\textbf{v}')\right)}\\
\end{aligned}
\end{equation}

\noindent Therefore, we construct the new input $u_{t}$ to each LSTM unit by concatenating the original word embedding and the attention vector in Equation  \ref{eq:att4} to distribute the senti-emoji information to each step. This model is called Multi-level Attention-based LSTM with Bi-sense Emoji Embedding (\textbf{MATT-BiE-LSTM}). We choose the same binary cross-entropy as the loss function with the same network configuration with WATT-BiE-LSTM.

\begin{equation}
\label{eq:att4}
\begin{aligned}
\textbf{u}&=[\textbf{w}_{t},\alpha_{t}'\cdot\textbf{v}']
\end{aligned}
\end{equation}

\begin{table*}[h]
	\centering
	\caption{Top-10 Most Frequently Used Emojis.}
	\label{tb:exp:data}
	\begin{tabular}{c|c|ccc|ccc}
		\hline
		\multirow{2}{*}{Ranking} & \multirow{2}{*}{Emoji} & \multicolumn{3}{c|}{AA-Sentiment} & \multicolumn{3}{c}{HA-Sentiment} \\ \cline{3-8} 
		&  & Positive & Negative & Pos-Ratio & Positive & Negative & Pos-Ratio \\ \hline
		1 & \includegraphics[height=0.7em]{38} & 164,677 & 62,816 & 0.724 & 333 & 162 & 0.673 \\
		2 & \includegraphics[height=0.7em]{53} & 146,498 & 4,715 & 0.969 & 184 & 32 & 0.852 \\
		3 & \includegraphics[height=0.7em]{50} & 105,329 & 4,566 & 0.958 & 181 & 23 & 0.887 \\
		4 & \includegraphics[height=0.7em]{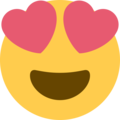} & 66,905 & 2,678 & 0.962 & 194 & 7 & 0.965 \\
		5 & \includegraphics[height=0.7em]{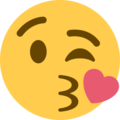} & 62,369 & 2,155 & 0.967 & 93 & 7 & 0.930 \\
		6 & \includegraphics[height=0.7em]{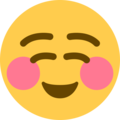} & 53,913 & 2,079 & 0.963 & 101 & 13 & 0.886 \\
		7 & \includegraphics[height=0.7em]{22} & 31,077 & 24,519 & 0.559 & 56 & 177 & 0.240 \\
		8 & \includegraphics[height=0.7em]{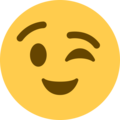} & 42,543 & 4,212 & 0.910 & 128 & 19 & 0.871 \\
		9 & \includegraphics[height=0.7em]{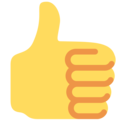} & 42,919 & 3,182 & 0.931 & 91 & 25 & 0.784 \\
		10 & \includegraphics[height=0.7em]{54} & 38,316 & 4,514 & 0.895 & 112 & 24 & 0.824 \\ \hline
	\end{tabular}
\end{table*}


\section{Experiments}
\subsection{Data Collection and Annotation}
\label{sec:exp:data}
\textbf{Data Collection}

\noindent We construct our own Twitter sentiment dataset by crawling tweets through the REST API \footnote{https://developer.twitter.com/en/docs} which consists of 350,000 users and is magnitude larger comparing to previous work. We collect up to 3,200 tweets from each user and follow the standard tweet preprocessing procedures to remove the tweets without emojis and tweets containing less than ten words, and contents including the urls, mentions, and emails. 

\noindent\textbf{Data Annotation}

\noindent For acquiring the sentiment annotations, we first use \emph{Vader} which is a rule-based sentiment analysis algorithm \cite{HuttoG14} for text tweets only to generate weak sentiment labels. The algorithm outputs sentiment scores ranging from -1 (negative) to 1 (positive) with neutral in the middle. We consider the sentiment analysis as a binary classification problem (positive sentiment and negative sentiment), we filter out samples with weak prediction scores within $\left(-0.60,0.60\right)$ and keep the tweets with strong sentiment signals. Emoji occurrences are calculated separately for positive tweets and negative tweets, and threshold is set to 2,000 to further filter out emojis which are less frequently used in at least one type of sentimental text. In the end, we have constructed a dataset with 1,492,065 tweets and 55 frequently used emojis in total. For the tweets with an absolute sentiment score over 0.70, we keep the auto-generated sentiment label as ground truth because the automatic annotation is reliable with high sentiment scores. On the other hand, we select a subset of the tweets with absolute sentiment scores between $\left(0.60,0.70\right)$ for manual labeling by randomly sampling, following the distribution of emoji occurrences where each tweet is labeled by two graduate students. Tweets are discarded if the two annotations disagree with each other or they are labeled as neutral. In the end, we have obtained 4,183 manually labeled tweets among which 60\% are used for fine-tuning and 40\% are used for testing purposes. The remainder of the tweets with automatic annotations are divided into three sets: 60\% are used for pre-training the bi-sense and conventional emoji embedding, 10\% for validation and 30\% are for testing. We do not include a ``neutral'' class because it is difficult to obtain valid neutral samples. For auto-generated labels, the neutrals are the samples with low absolute confidence scores and their sentiments are more likely to be model failures other than ``true neutrals''. Moreover, based on the human annotations, most of the tweets with emojis convey non-neutral sentiment and only few neutral samples are observed during the manual labeling which are excluded from the manually labeled subset.

In order to valid our motivation that emojis are also extensively used in tweets that contain contradictory information to the emoji sentiments, we calculate the emoji usage in Table \ref{tb:exp:data} according to the sentiment labels where \emph{Pos-Ratio} means the percentage of each emoji occurs in the positive tweets over its total number of occurrences, \emph{AA} and \emph{HA} indicate automatic-annotation and human-annotation, respectively. We present the top-10 most frequently used emojis in our dataset and observe a slight difference in the Pos-Ratios between AA and HA dataset because of the randomness involved in the sampling process. Results from both of the datasets show a fair amount of emoji use in both positive and negative tweets. For example, it is interesting to notice that emoji (\includegraphics[height=0.7em]{22}) occurs more in the positive tweets in with the automatic annotations, while emojis with strong positive sentiment have also been used in negative tweets with about 5\% occurrences, such as (\includegraphics[height=0.7em]{53}, \includegraphics[height=0.7em]{50}, and \includegraphics[height=0.7em]{39}). Given the averaged positive ratio among all emojis in the whole dataset is about 74\% and that most emojis have been extensively used in tweets containing both positive and negative sentiments, it suggests that distinguishing the emoji occurrences in both sentiments via bi-sense embedding is worth investigating. Additionally, we observe the \emph{Pos-Ratios} of the AA-sentiment and HA-sentiment have little differences which are due to two main reasons: 1) Some tweets we sampled to construct the HA-sentiment are discarded because the annotators have disagreements and we only keep the samples that we are confident about; 2) Tweets with absolute sentiment scores between (0.60,0.70) are selected for manual labeling as discussed in Section \ref{sec:exp:data}, which are lower than the tweets used to construct the AA-sentiment (0.7 and above). The lower sentiment scores indicate that Vader is less reliable on the samples of HA-sentiment dataset and the sentiments of these tweets are more likely to be affected by emojis.

\begin{table*}[t]
	\centering
	\caption{Twitter Sentiment Analysis.}
	\label{tb:exp:acc}
	\begin{tabular}{c|ccccc|ccccc}
		\hline
		\multirow{2}{*}{\textbf{Models}} & \multicolumn{5}{c|}{\textbf{AA-Sentiment}} & \multicolumn{5}{c}{\textbf{HA-Sentiment}} \\ \cline{2-11} 
		& Precision & Recall & ROC Area & Accuracy & F1 Score & Precision & Recall & ROC Area & Accuracy & F1 Score \\ \hline
		T-LSTM & 0.921 & \textbf{0.901} & 0.931 & 0.866 & 0.911 & 0.708 & 0.825 & 0.774 & 0.707 & 0.762 \\
		E-LSTM & 0.944 & 0.86 & 0.933 & 0.855 & 0.900 & 0.816 & 0.825 & 0.855 & 0.794 & 0.820 \\
		ATT-E-LSTM & 0.948 & 0.890 & 0.954 & 0.879 & 0.918 & 0.825 & 0.868 & 0.878 & 0.820 & 0.846 \\ \hline
		BiE-LSTM & 0.961 & 0.891 & 0.966 & 0.890 & 0.924 & 0.822 & 0.881 & 0.898 & 0.824 & 0.850 \\
		MATT-BiE-LSTM & \textbf{0.972} & 0.895 & \textbf{0.975} & \textbf{0.900} & \textbf{0.932} & \textbf{0.831} & 0.872 & 0.890 & 0.826 & 0.851 \\
		WATT-BiE-LSTM & 0.949 & 0.895 & 0.960 & 0.883 & 0.921 & 0.830 & \textbf{0.889} & \textbf{0.899} & \textbf{0.834} & \textbf{0.859} \\ \hline
	\end{tabular}
\end{table*}

\subsection{Sentiment Analysis}
\noindent\textbf{Models}\newline
\noindent We set up the baselines and proposed models as follows:

\noindent \emph{LSTM with text embedding}: CNNs and LSTMs are widely used to encode textual contents for sentiment analysis in \cite{cliche2017bb_twtr,vo2017multi} and many online tutorials. Here we select the standard LSTM with pre-trained word embedding as input, and add one fully-connected layer with \emph{sigmoid} activation top of the LSTM encoder (same as all other models), denoted as \textbf{T-LSTM}.

\noindent \emph{LSTM with emoji embedding}: We consider the emoji as one special word and input both pre-trained text and emoji embeddings into the same LSTM network, namely \textbf{E-LSTM}. Similarly, we concatenate the pre-trained bi-sense emoji embedding as one special word to feed into the LSTM network. This model is called \textbf{BiE-LSTM}.

\noindent \emph{Attention-based LSTM with emojis}:We also use the word-emoji embedding to calculate the emoji-word attention following Equation \ref{eq:att3} and \ref{eq:att4}, and the only difference is that we replace the attention-derived senti-emoji embedding with the pre-trained word-emoji embedding by \emph{fasttext}, denoted as \textbf{ATT-E-LSTM}.

\noindent \emph{LSTM with bi-sense emoji embedding (proposed)}: As we have introduced in Section \ref{sec:attlstm}, we propose two attention-based LSTM networks based on bi-sense emoji embedding, denoted as \textbf{MATT-BiE-LSTM} and \textbf{WATT-BiE-LSTM}.

\vspace*{0.8mm}
\noindent\textbf{Evaluation}\newline
\noindent We evaluate the baseline and proposed models on sentiment analysis by F1 scores and accuracies based on the auto-annotated testing set (AA-Sentiment) and human-annotated testing set (HA-Sentiment), as shown in Table \ref{tb:exp:acc}. We only test the models after fine-tuning with a subset of the samples with human annotations because training exclusively on the samples with auto-generated weak labels results in relatively poor performances when tested with human annotated data indicating the models after fine-tuning are more robust. The F1 scores and accuracies are overall higher with the AA-Sentiment than the results with HA-sentiment, indicating that the HA-Sentiment is a more challenging task and the sentiments involved are more difficult to identify supported by their relatively lower sentiment scores returned from \emph{Vader}. We still, however, observe competitive results from HA-Sentiment showing that the models are well-trained and robust to noisy labels with the help of fine-tuning with human annotated data. The T-LSTM baseline achieves decent performance in both experiments with accuracies of 86.6\% and 70.7\% showing that LSTM is an effective encoder for sentiment analysis as suggested by the references. The models with proposed bi-sense emoji embedding obtain accuracies over 82.4\% and we observe improvements on the performance with the attention-based LSTM from our proposed model MATT-BiE-LSTM and WATT-BiE-LSTM, which is consistent with that ATT-E-LSTM (F1@84.6\%, accuracy@82.0\% on HA-Sentiment) outperforms significantly T-LSTM and E-LSTM.

\textbf{Emoji information is useful in sentiment analysis}. Most models outperforms the baseline T-LSTM in both dataset suggesting that the emoji information is useful for sentiment analysis as a complement to the textual contents, even with the naive use of emoji embeddings (E-LSTM) when tested with HA-Sentiment. We observe that E-LSTM obtains similar performance to T-LSTM with AA-Sentiment but a significant gain over the T-LSTM when tested with HA-Sentiment indicating that sentiment information is helpful and necessary when the hidden sentiment is relatively subtle and the task is more challenging. 

\textbf{Bi-sense emoji embedding helps}. All the models using bi-sense emoji embedding perform significantly better than the baseline models without emoji feature or with word-emoji embedding. BiE-LSTM outperforms T-LSTM and E-LSTM significantly with the same utilization of emoji embedding indicates that the proposed bi-sense emoji embedding is capable of extracting more informative and distinguishable vectors over the use of conventional word embedding algorithms, which is consistent based on the comparisons between the proposed models (MATT-BiE-LSTM and WATT-BiE-LSTM) with bi-sense emoji embedding and the baseline model ATT-E-LSTM with word-emoji embedding and attention. 

\textbf{Attention mechanism aligns and performs well with bi-sense embedding}. MATT-BiE-LSTM and WATT-BiE-LSTM obtain similar performances when tested on both \emph{Vader} and human annotated samples, though their ways of computing the attention (weights and vectors) are different that WATT computes attention weights and the senti-emoji embeddings guided by each word, and MATT obtains the senti-emoji embedding based on the LSTM encoder on the whole contexts and computes the attention weights of the senti-emoji embedding across all words. Both models outperforms the state-of-the-art baseline models including ATT-E-LSTM. The proposed attention-based LSTM can be further extended to handle tasks involving multi-sense embedding as inputs, such as the word-sense embedding in NLP, by using context-guide attention to self-select how much to attend on each sense of the embeddings each of which correspond to a distinct sense of semantics or sentiments. In this way we are able to take advantage of the more robust and fine-grained embeddings.

\begin{figure}[t]
	\subfloat[OMG no pressure, I'll be happy to hang out with whichever lovely people are there.]{%
		\hbox{\hspace{-0.72em}}\includegraphics[clip,width=1.045 \columnwidth]{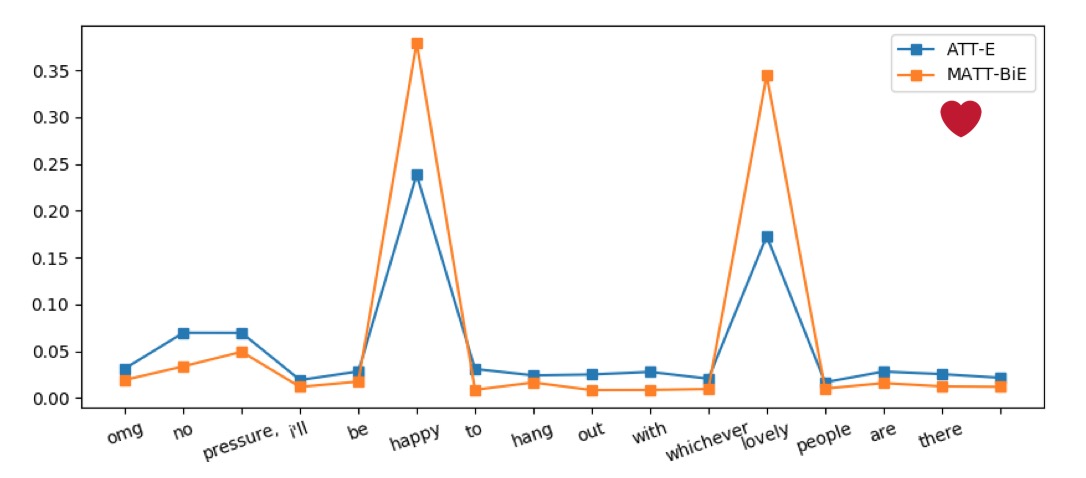}%
	}
	
	\subfloat[This feels both ridiculous \& desperate. I'd pay \$100 max for a hoodie. Am I cheap?]{%
		\hbox{\hspace{-0.15em}}\includegraphics[clip,width=1.01\columnwidth]{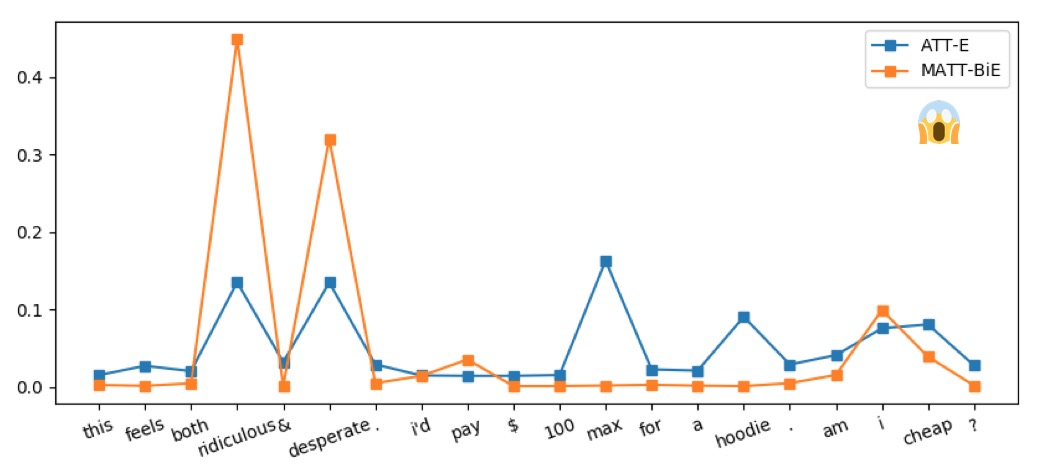}%
	}
	
	\subfloat[Amazing! Can't wait to read this after I'm finished with the first.]{%
		\hbox{\hspace{-0.35em}}\includegraphics[clip,width=1.0\columnwidth]{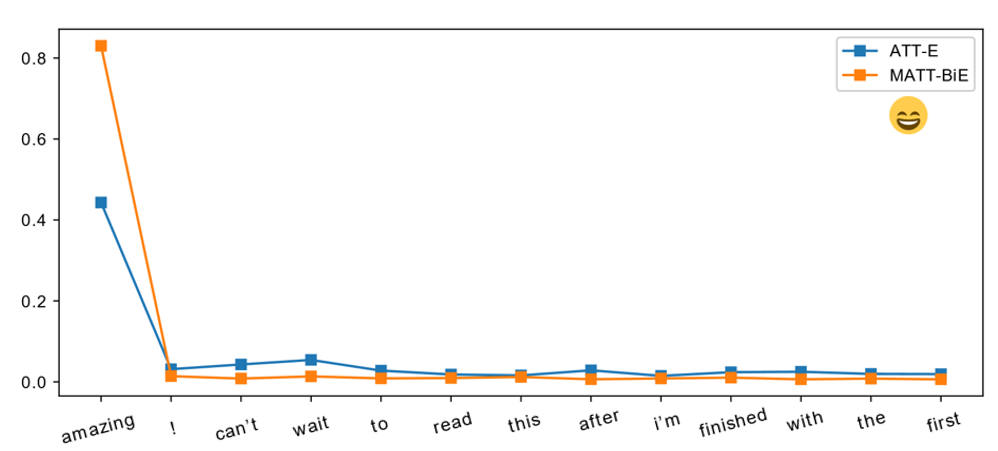}%
	}
	\caption{Attention weights obtained by senti-emoji embedding and word-emoji embedding across words. Tweet contexts are given in sub-captions.}
	\label{fig:exp:eg}
\end{figure}

\begin{figure}[b]
	\subfloat[Bi-sense Emoji Embedding]{%
		\includegraphics[clip,width=0.25\textwidth]{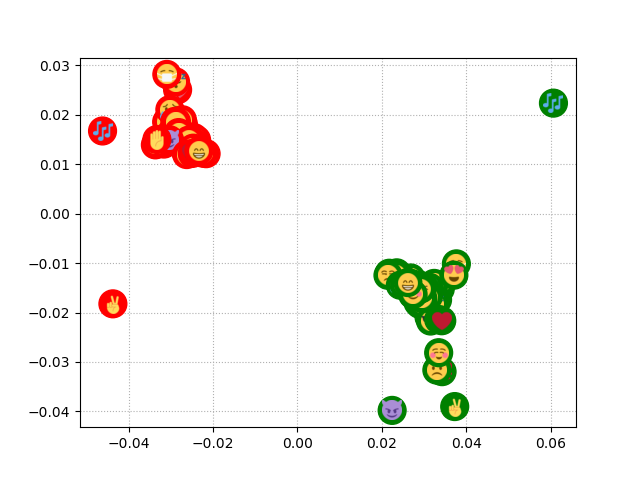}%
	}
	\subfloat[Positive-sense Embedding]{%
		\includegraphics[clip,width=0.25\textwidth]{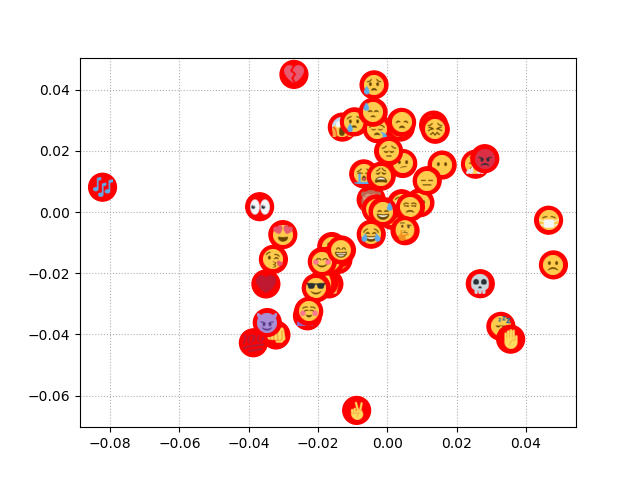}%
	}
	
	\subfloat[Negative-sense Embedding]{%
		\includegraphics[clip,width=0.25\textwidth]{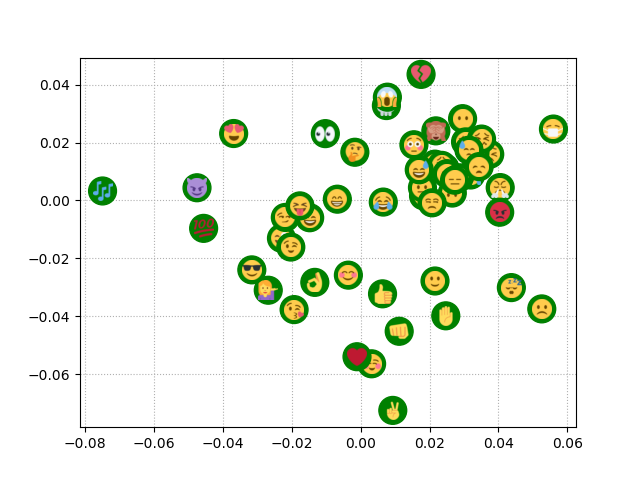}%
	}
	\subfloat[Positive-sense $-$ Negative-sense]{%
		\includegraphics[clip,width=0.25\textwidth]{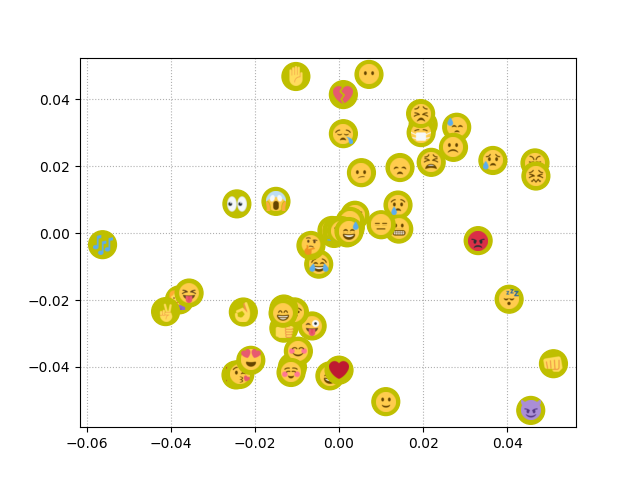}%
	}
	\caption{\emph{t-SNE} visualization of bi-sense emoji embedding. Positive-sense embeddings are paired with red circles, negative-sense embeddings are paired with green circles, and their subtractions are paired with yellow circles, respectively. Best viewed when zoomed in.}
	\label{fig:exp:vis}
\end{figure}

\subsection{Qualitative Analysis}
In order to obtain insights about why the more fine-grained bi-sense emoji embedding helps in understanding the complexed sentiments behind tweets, we visualize the attention weights for ATT-E-LSTM and MATT-BiE-LSTM for comparison. The example tweets with corresponding attention weights calculated by word-emoji embedding and senti-emoji embedding are shown in Figure \ref{fig:exp:eg}, where the contexts are presented in the captions. The emojis used are \includegraphics[height=0.8em]{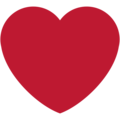}, \includegraphics[height=0.8em]{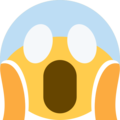}, and \includegraphics[height=0.8em]{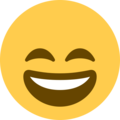}, respectively.

In Figure \ref{fig:exp:eg}(a), the ATT-E-LSTM model (baseline) assigns relatively more weights on the word ``no'' and ``pressure'', while MATT-BiE-LSTM attends mostly on the word ``happy'' and ``lovely''. The different attention distributions suggest that the proposed senti-emoji embedding is capable of recognizing words with strong sentiments that are closely related to the true sentiment even with the presence of words with conflicting sentiments, such as ``pressure'' and ``happy''. while ATT-E-LSTM tends to pick up all sentimental words which could raise confusions. The senti-emoji embedding is capable of extracting representations of complexed semantics and sentiments which help guide the attentions even in cases when the word sentiment and emoji sentiment are somewhat contradictory to each other. From Figure \ref{fig:exp:eg}(b) and (c) we can observe that the ATT-E-LSTM assigns more weights on the sentiment-irrelevant words than the MATT-BiE-LSTM such as ``hoodies'', ``wait'' and ``after'', indicating that the proposed model is more robust to irrelevant words and concentrates better on important words. Because of the senti-emoji embedding obtained through bi-sense emoji embedding and the sentence-level LSTM encoding on the text input (described in Section \ref{sec:attlstm}), we are able to construct a more robust embedding based on the semantic and sentiment information from the whole context compared to the word-emoji embedding used in ATT-E-LSTM which takes only word-level information into account. 

\subsection{Bi-sense Emoji Embedding Visualization}
To gain further insights on the bi-sense emoji embedding, we use \emph{t-SNE} \cite{maaten2008visualizing} to project high-dimensional bi-sense embedding vectors into a two-dimensional space and preserving the relative distances between the embedding vectors at the same time. In Figure \ref{fig:exp:vis} we visualize the bi-sense emoji embedding, positive-sense embedding, negative-sense embedding and the subtraction between positive and negative sense embeddings of each emoji, respectively. The subtraction of an emoji between its two sense embeddings indicates the semantic differences between emoji usages in positive and negative sentimental contexts, similarly to the objective of word embeddings \cite{mikolov2013distributed}. The positive-sense of emoji (\includegraphics[height=0.8em]{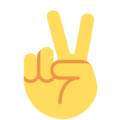} and \includegraphics[height=0.8em]{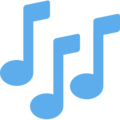}), and the negative-sense of emoji (\includegraphics[height=0.8em]{44.png}, \includegraphics[height=0.8em]{8.png} and \includegraphics[height=0.8em]{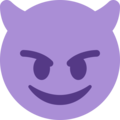}) are embedded far from the two main clusters as observed in Figure \ref{fig:exp:vis}(a), suggesting that the semantics of these emojis are different from the other popular emojis. The positive-sense embedding and negative-sense embeddings are clustered well with no intersection with each other. Such observation supports our objective of applying bi-sense emoji embedding because there exist such significant differences in the semantics of each emoji when appears in positive and negative sentimental contexts, and it is well-motivated to consider the emoji usages individually according to the sentiment of the contexts to extract the more fine-grained bi-sense embedding. Additionally, we observe consistent patterns in the Figure \ref{fig:exp:vis}(b), (c) and (d) where the sentiments conveyed in the emojis become an important factor. For example, emojis with positive sentiments such as (\includegraphics[height=0.8em]{54.png}, \includegraphics[height=0.8em]{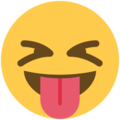} and \includegraphics[height=0.8em]{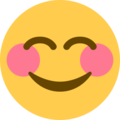}), and emojis with negative sentiment such as (\includegraphics[height=0.8em]{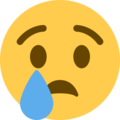}, \includegraphics[height=0.8em]{22.png} and \includegraphics[height=0.8em]{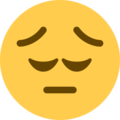}) are embedded into one clusters in both positive-sense and negative-sense space. The embedding subtractions of emojis in Figure \ref{fig:exp:vis}(d) shows the different usages of emojis across sentiments are similar between emojis and preserve the cluster patterns observed in Figure \ref{fig:exp:vis} (b) and (c).

\section{Conclusions}
In this paper, we present a novel approach to the task of sentiment analysis and achieve the state-of-the-art performance. Different from the previous work, our method combines a more robust and fine-grained bi-sense emoji embedding that effectively represents complex semantic and sentiment information, with attention-based LSTM networks that selectively attend on the correlated sense of the emoji embeddings, and seamlessly fuse the obtained senti-emoji embeddings with the word embeddings for a better understanding of the rich semantics and sentiments involved. In the future, we plan to further extend our attention-based LSTM with bi-embedding work frame to tackle tasks involving multi-sense embedding such as the learning and applications of word-sense embedding.

\section*{Acknowledgement}
We would like to thank the support of New York State through the Goergen Institute for Data Science, and NSF Award \#1704309.

\bibliographystyle{ACM-Reference-Format}
\balance
\bibliography{sample-bibliography}

\end{document}